\documentclass[sigconf]{acmart}

\setcopyright{acmcopyright}
\copyrightyear{2023}
\acmYear{2023}
\acmDOI{XXXXXXX.XXXXXXX}

\acmConference[BDCAT '23]{The 10th IEEE/ACM International Conference on Big Data Computing, Applications and Technologies}{December 04--07,
  2023}{Messina, Italy}
\usepackage{csquotes}
\usepackage[inline]{enumitem}
\usepackage{graphicx}
\usepackage{subfig}
\usepackage{tcolorbox}
\usepackage{xspace}
\usepackage{hyperref}

\newcommand{\nathaniel}[1]{{\color{Blue} \textbf{Nathaniel:} \enquote{#1}}}

\newcommand{\greg}[1]{{\color{Fuchsia} \textbf{Greg:} \enquote{#1}}}

\newcommand{\cmp}{commercial model provider\xspace}
\newcommand{\cmps}{commercial model providers\xspace}

\newcommand{\lsms}{large-scale models\xspace}

\newcommand{\tpm}{TPM\xspace}
\newcommand{\tpms}{TPMs\xspace}

\begin{document}
\title{
    Trillion Parameter AI Serving Infrastructure for Scientific Discovery: A Survey and Vision
}


\author[N. Hudson]{Nathaniel Hudson}
\affiliation{
    \institution{University of Chicago}
    \institution{Argonne National Laboratory}
    \city{}
    \country{}
}

\author[J. G. Pauloski]{J. Gregory Pauloski}
\affiliation{
    \institution{University of Chicago}
    \city{}
    \country{}
}

\author[M. Baughman]{Matt Baughman}
\affiliation{
    \institution{University of Chicago}
    \city{}
    \country{}
}

\author[A. Kamatar]{Alok Kamatar}
\affiliation{
    \institution{University of Chicago}
    \city{}
    \country{}
}

\author[M. Sakarvadia]{Mansi Sakarvadia}
\affiliation{
    \institution{University of Chicago}
    \city{}
    \country{}
}

\author[L. Ward]{Logan Ward}
\affiliation{
    \institution{Argonne National Laboratory}
    \city{}
    \country{}
}

\author[R. Chard]{Ryan Chard}
\affiliation{
    \institution{Argonne National Laboratory}
    \city{}
    \country{}
}

\author[A. Bauer]{Andr\'{e} Bauer}
\affiliation{
    \institution{University of Chicago}
    \institution{Argonne National Laboratory}
    \city{}
    \country{}
}

\author[M. Levental]{Maksim Levental}
\affiliation{
    \institution{University of Chicago}
    \city{}
    \country{}
}

\author[W. Wang]{Wenyi Wang}
\affiliation{
    \institution{University of Chicago}
    \city{}
    \country{}
}

\author[W. Engler]{Will Engler}
\affiliation{
    \institution{University of Chicago}
    \city{}
    \country{}
}

\author[O. P. Skelly]{Owen Price Skelly}
\affiliation{
    \institution{University of Chicago}
    \city{}
    \country{}
}

\author[B. Blaiszik]{Ben Blaiszik}
\affiliation{
    \institution{University of Chicago}
    \city{}
    \country{}
}

\author[R. Stevens]{Rick Stevens}
\affiliation{
    \institution{University of Chicago}
    \institution{Argonne National Laboratory}
    \city{}
    \country{}
}

\author[K. Chard]{Kyle Chard}
\affiliation{
    \institution{University of Chicago}
    \institution{Argonne National Laboratory}
    \city{}
    \country{}
}

\author[I. Foster]{Ian Foster}
\affiliation{
    \institution{University of Chicago}
    \institution{Argonne National Laboratory}
    \city{}
    \country{}
}

\renewcommand{\shortauthors}{Hudson et al.}


\begin{abstract}
    Deep learning methods are transforming research, enabling new techniques, and ultimately leading to new discoveries.
    As the demand for more capable AI models continues to grow, we are now entering an era of Trillion Parameter Models (\tpm{}), or models with more than a trillion parameters---such as 
    Huawei's PanGu-$\Sigma$.
    We describe a vision for the ecosystem of \tpm{} users and providers that caters to the specific needs of the scientific community.
    We then outline the significant technical challenges and open problems in system design for serving \tpms{} to enable scientific research and discovery. Specifically, we describe the requirements of a comprehensive software stack and interfaces to support the diverse and flexible requirements of researchers.
\end{abstract}



\maketitle

\section{Introduction}
\label{sec:intro}

\begin{figure*}
    \centering
    \subfloat[AI Model Size over Time\label{fig:model_benchmark}]{
        \includegraphics[width=0.425\linewidth]{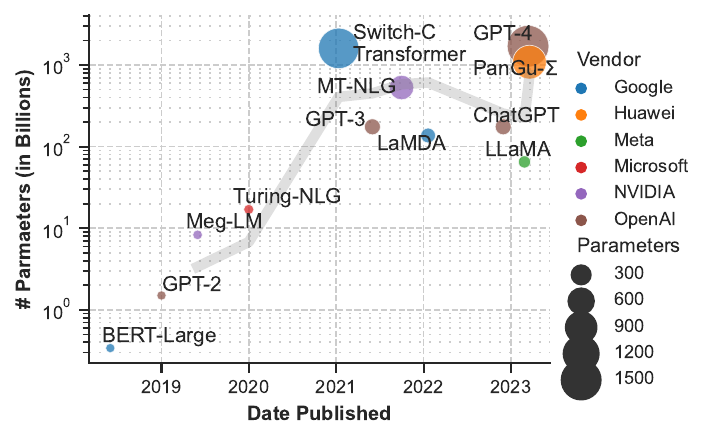}
    }
    \hfill
    \subfloat[System Benchmarks over Time\label{fig:system_benchmark}]{
        \includegraphics[width=0.425\linewidth]{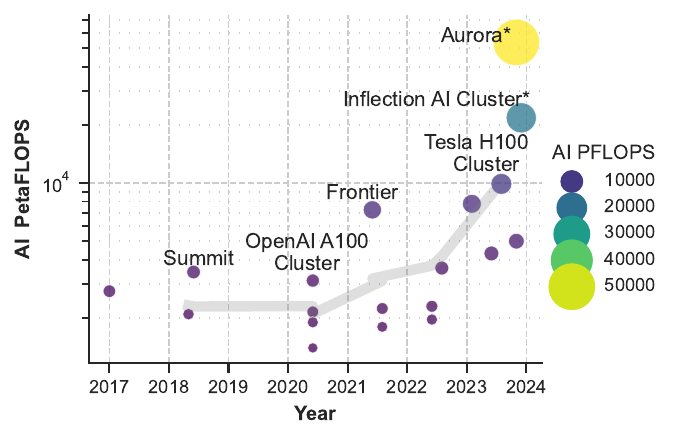}
    }
    \caption{
        (\ref{fig:model_benchmark}) Size of AI language models over the past few years. The gray line is the trend-line as a rolling average. The top-right corner features 3 TPMs: 
        Switch-C Transformer~\cite{fedus2022switch}, PanGu-$\Sigma$~\cite{ren2023pangusigma}, and (rumored) GPT-4~\cite{gpt4}.
        (\ref{fig:system_benchmark}) The benchmarked AI-PetaFLOPS of state-of-the-art HPC systems over time. Note: An asterisk means the \textit{anticipated} benchmark numbers.
    }
    \label{fig:benchmarks}
\end{figure*}

In recent years, much attention has been drawn to \textit{Artificial Intelligence}~(AI) and, more specifically, the subfield of \textit{Deep Learning}~(DL). DL uses deep neural networks. These models, if given sufficient data, can learn very complex patterns hidden in large amounts of data.  
The success of DL for a variety of different problems has led to the release of various public-facing models, such as OpenAI's ChatGPT~\cite{chatgpt} and DALL-E~\cite{dalle}, Google's Bard~\cite{bard} and LaMDA~\cite{thoppilan2022lamda}, and Huawei's PanGu-$\Sigma$~\cite{ren2023pangusigma}. 
Notably, ChatGPT has sparked discussion about new applications of AI due to its ability to generate textual answers to user provided questions, 
write songs in the style of a pop music artist, and draft professional emails.
Even before the proliferation of \textit{Large Language Models}~(LLMs) like ChatGPT, AI models like AlphaFold~\cite{jumper2021alphafold} and AlphaGo~\cite{silver2016alphago} demonstrated utility in specific areas.
Following these recent advancements, many scientists are now using AI as a tool in their research workflows~\cite{humbird2020fusion,radivojevic2020biology,hong2021moleculesnlp,zvyagin2022genslm,ward2021colmena}. 

Looking towards the future of AI for science, it will become increasingly important to consider what is required to train and serve AI models for science. As we enter the era of \textit{Trillion Parameter Models}~(\tpms{})---models with at least one trillion parameters---we must also consider the specific challenges of serving large models with significant resource needs. Huawei's PanGu-$\Sigma$~\cite{ren2023pangusigma} is an example of a TPM. 
First, at least one large-scale \textit{High-Performance Computing}~(HPC) system is needed to train and serve such a model with sufficient throughput. Second, one would need to exhaustively curate and collect enormous troves of scientific data to train a \tpm{} for scientific purposes. This will require active collaboration with scientific researchers across a large number of disciplines. Third, facilities with the technical talent and resources necessary to maintain the model and data infrastructure are essential to ensure that \tpm{}s are accessible. 

These three aforementioned challenges are general and apply to any system which builds and hosts a \tpm{}. Scientific researchers, on the other hand, have unique requirements (e.g., more fine-grained access to the \tpm{}) that demand custom solutions compared to those used to serve commercial AI models to the public.
For small models that can fit on a single machine, this is a relatively simple problem. However, for a \tpm{}, the problem becomes more challenging. 
There remain open problems about designing friendly users interfaces that are still comprehensive enough to meet the flexible needs of scientists.
Supporting these requirements at the largest scales, e.g., exascale computing projects~\cite{paul2017ecp}, will introduce new system design challenges to support \tpm{}s for scientific research.

This paper presents a vision and the associated technical challenges with building systems for serving \tpm{}s for scientific applications. The focus of this paper is \textit{not} training \tpm{}s. 
Section~\ref{sec:history} presents an overview of the history of serving \lsms{}. 
Section~\ref{sec:3_landscape} discusses the current landscape of research into \lsms{} and the application of AI in scientific workflows. 
Section~\ref{sec:vision} presents our vision for incorporating \tpm{}s into scientific research.
Section~\ref{sec:challenges} explores several challenges associated with implementing such a system.
Section~\ref{sec:conclusions} concludes our vision.

\section{History of Serving Large Models}
\label{sec:history}
Starting in the 2010s, significant advancements in both machine learning algorithms and computer hardware ushered in more efficient techniques for training deep neural networks. Consequently, businesses began integrating ML and DL models into their daily operations. However, a challenge became evident as  managing the ML lifecycle proved to be a slow and challenging process when it came to scaling for business applications. In response to this need for more efficient management, the concept of MLOps emerged in 2015. Often referred to as ``DevOps for machine learning,'' MLOps streamlines the entire process, from development to deployment, by incorporating essential elements like monitoring, validation, and governance into the management of ML models. 

IBM Watson took the pioneering step of becoming the first AI-as-a-Service platform in 2013. This led to other companies such as Amazon following suit. In the subsequent years, a plethora of model repositories and serving systems have been introduced, each contributing to the evolving landscape of ML and AI services. In the following sections, we delve into some of these notable services.

Model repositories, such as HuggingFace Transformers~\cite{huggingface}, ModelHub~\cite{modelhub}, and Caffe Zoo~\cite{caffe}, to name a few, play a pivotal role in the ML lifecycle by providing tools to facilitate consistent, reproducibile management and publication of models. 
Model repositories provide tools to systematically track, version, manage, disseminate, and compare different iterations of models. Repositories also promote collaboration by allowing teams to share models, documentation, and associated metadata, streamlining the process of model deployment and integration. 

Model serving platforms provide capabilities to host trained models for rapid inference. Many serving platforms also include functionality to manage the ML lifecycle. 
Cloud-hosted MLOps solutions, such as  Amazon SageMaker~\cite{awssagemaker}, 
provide flexible scalability via on-demand computational resources that can be dynamically adjusted to meet requirements, be it data preprocessing, model training, or deployment. These services offer a myriad of tools tailored for different stages of the ML lifecycle, from data storage and \textit{Extract-Transform-Load}~(ETL) operations to hyperparameter tuning and model serving. This cohesive ecosystem promotes streamlined workflows, simplifies experimentation, and accelerates the time-to-market for ML solutions. Managed services include built-in monitoring, logging, and security protocols. 

The practice of implementing MLOps on-premises, carries significant importance in the context of data management, security, and operational flexibility while maintaining data and model sovereignty. Local MLOps, such as MLFlow~\cite{zaharia2018accelerating}, Kubeflow~\cite{kubeflow}, and Clipper~\cite{clipper}, offer more customized infrastructure solutions, enabling users to tailor their computational resources and networking based on specific project requirements, thereby avoiding potential bottlenecks or inefficiencies that could arise in general cloud environments. This bespoke approach can lead to optimized performance and costs while reducing latency.

Utilizing HPC resources for hosting models offers several distinct advantages over traditional cloud platforms by providing tuned infrastructure designed to accommodate ML workloads and capabilities to host and share large data and models that would otherwise be impractical. Ray Serve~\cite{ray}, DLHub~\cite{dlhub}, and Garden~\cite{garden} are three projects that enable the use of HPC resources, exposing the benefits from direct access to large, high-speed storage systems and tailored network configurations, enhancing data transfer rates and reducing latency. 

\section{Artificial Intelligence in Science}
\label{sec:3_landscape}
Scientists and researchers are broadly interested in \lsms{} from two angles: 
\begin{enumerate*}[label=\textit{(\roman*)}]
    \item studying the \lsms{} directly 
    and 
    \item using \lsms{} as a research tool.
\end{enumerate*}
While the focus of this work is to support \tpms{} as a tool for scientific applications and workflows, we argue that studying and understanding a \tpm{} is a necessary component, or even prerequisite, to using the \tpm{} for science.
For example, Meta's Galactica~\cite{taylor2022galactica}, a large language model for science, was quickly removed due the models propensity to generate fake or misleading scientific information~\cite{mit2022galactica}.
Science-oriented \tpms{} necessitate more rigorous evaluation, and thus, we first discuss work in studying AI models before describing the landscape of AI for science.

\subsection{Studying AI Models}
\label{sec:studying_ai_models}

The models used in modern AI (e.g., DNNs) 
are made up of several components that enable them to learn very complex functions that map inputs to outputs. These learned functions are often so obscure that a human  struggles to make sense of it. To this end, expansive research is conducted with the goal of better understanding these models. 

A lot of AI research focuses on studying how to create better deep learning models from different perspectives and approaches. The de facto approach to implementing models that achieve greater accuracy has been to increase the model size (i.e., parameters)~\cite{kaplan2020scaling}. 
Another approach focuses on designing new mathematical structures to include in a DNN to capture relationships in data, for example long short-term memory~\cite{hochreiter1997lstm}, gated recurrent units~\cite{cho-etal-2014-learning}, and transformers~\cite{vaswani2017attention}. 
Finally, others have instead taken the approach of trying to find the most optimal sequence of DNN components to maximize model accuracy~\cite{luo2018neural, benardos2007optimizing, idrissi2016genetic}. In these cases, researchers have applied strong mathematical reasoning, apply heuristics, or use AI-based techniques to tackle the problem (e.g., neural architecture search~\cite{ren2021comprehensive, liu2018progressive}). 
A subset of these approaches will also consider constraints to avoid just carelessly increasing the model size~\cite{lu2019neural}. This research is attractive to those that work in resource-constrained settings in the computing continuum (e.g., edge and fog computing systems~\cite{hudson2021qos, zhao2020improving}). 

Compared to classical statistical methods (e.g., linear regression), DNNs are often very large and complex. While DNNs can learn over complex data without requiring rigid feature engineering, this makes them very difficult to directly interpret and understand. The ability to interpret a trained DNN is not necessary for all end users. However, it is an open problem in AI for several reasons. One relevant reason is to identify and eliminate bias from pre-trained models. The concern around bias in models has grown as generative models (e.g., ChatGPT) become widely accessible. 
A second reason is to better understand points of failure for AI models responsible for automated decision-making. Imagine a scenario where a self-driving car operated by AI is involved in a crash. In this case, it is crucial to understand how the model reached a decision to take the actions it did leading up to the crash to have a more informed idea of who is at fault. A third  example is to simply better understand the training data. In classical statistical methods, much can be learned about the nature of the data by using regression. A researcher may be able to discover phenomenon based on how the model was fit to their data. This is more challenging with DNNs because of their complexity. If a DNN is trained on a large amount of scientific data, interpretability may lead us to groundbreaking scientific discoveries that may not be detected by manually interpreting the data.

Finally, much research in deep learning has focused on expanding the modalities of data considered. 
\textit{Convolutional Neural Networks}~(e.g., ConvNets) are widely-used DNN architectures that are applied to several types of data (e.g., images, videos, sound represented as spectrograms). However, ConvNets are designed for Euclidean data (e.g., regular $n$-dimensional data), such as images or sequences. Many data are naturally non-Euclidean. Simple examples include the topologies of social networks, gene regulatory networks, Internet traffic, supply chains, and biological systems. Much research has focused on converting these non-Euclidean data into vectorized forms such that Euclidean DNN components can learn on them~\cite{narayanan2017graph2vec, grohe2020word2vec}. This, of course, has limitations because the full nature of the original data is not completely expressed. Increasingly, researchers are exploring new DNN architectures, such as Graph Neural Networks~\cite{ZHOU202057}, to natively operate on non-Euclidean data (e.g., graphs, point clouds). Research in this area could empower future scientific discoveries as we make progress on designing architectures that are naturally compatible with more modalities of data.

\subsection{AI for Science}
\label{sec:ai_for_science}
The steady adoption of AI within science~\cite{cao2023moformer,ward2021colmena,zvyagin2022genslm,ward2023colmena} has primed the rapid 
creation and realization of opportunities for deploying \tpms{}.
Many modern applications of AI evolved from needs which were once and still solved by 
conventional statistical or computational models, such as ``how will X happen under Y conditions'' 
(i.e., supervised learning) or ``which of Z are the most similar'' (unsupervised learning).
The approaches taken in AI to answer these scientific questions have diverged over time, from Bayesian-based methods to prompt engineering for modern-day LLMs.

A suitable vignette for the role and approaches of AI in science is the design of molecules.
Science in the 1960s approached the challenge in a very human-centric way through regression models with 
clear, understandable functional forms meant as \textit{tools for use by human scientists}~\cite{hansch1964qsar}.
These Quantitative Structure-Property Relationship (QSPR) models are used extensively today in a similar context:
tools meant for and primarily employed by humans~\cite{le2012qsar}.

Advances in computing and mathematics 
have opened a different modality for machine learning: tools smart enough to act as intelligent actors.
More advanced roles require solving more intelligent tasks, such as ``produce something that is similar to X'' (e.g., using an autoencoder~\cite{rgb2018molecularvae})
or ``design an experimental campaign to optimize Y'' (e.g., using active learning~\cite{warmuth2003activelearnchem}).
Human involvement in such tasks 
is less frequent, yet this has become an accepted practice with 
AI directing mobile robots in laboratories.

The role of AI becoming more advanced with time is hardly limited to molecular design nor even chemical data.
A Department of Energy report on \enquote{Advanced Research Directions on AI for Science, Energy, and Security} outlines six key high-level AI approaches~\cite{doe2022aiworkshop}: 
\begin{enumerate*}[label=\textit{(\roman*)}]
    \item AI and surrogate models for scientific computing; 
    \item AI foundation models for scientific knowledge discovery, integration, and synthesis; 
    \item AI for advanced property inference and inverse design; 
    \item AI-based design, prediction, and control of complex engineered system; 
    \item AI and robotics for autonomous discovery; 
    and 
    \item AI for programming and software engineering.
\end{enumerate*}

The next stage in AI development is for AIs to learn from and participate in science in the same way as humans.
The previously-described examples of AI were purpose-built from curated data and communicate with humans through
restricted software interfaces or bypass them completely by only communicating with robots through restrictive software interfaces.
Limited access to data excludes AI from learning from broader scientific discourse and the limited ways to interact with the human world reduce their potential impacts.
The increased ability of AIs to learn from data in its natural forms and perform increasingly-complex tasks mandate new ways for the AI to interact with the human world.

TPMs will open opportunities for using AI in science that, in particular, place intensive demands on computing systems. 
These opportunities include:
\begin{enumerate*}
    \item \textbf{Collaborative Literature Comprehension} where a human and an AI trained on scientific literature can collaboratively develop hypotheses based on the current literature. Responding to queries and summarizing existing literature requires near-real-time inference to be suitable partner for a scientist.
    \item \textbf{Multi-modal learning} is likely to become a prevalent technique as many types of scientific data are naturally presented in non-textual forms. The breadth of data types and consistent development of techniques to learn from scientific data will create constant needs to fine-tune new models. Once created, such models could be useful in demanding inference tasks such as on-demand data analysis in scientific and experiments workflows.
    \item \textbf{Re-configuring Computational Workflows and Laboratories} is the next step over the experimental design techniques of today which are bound to existing workflows. Writing new software or an experimental procedure is a complex task and may involve iteration with digital twins to evaluate the safety or effectiveness of a workflow, which means that longer response times are not as mandatory as in collaborative uses. However, it does mean that the software systems for TPM inference must be tied closely with compute capable of quickly evaluating the TPM's proposals.
\end{enumerate*}










\section{Vision for the Future}
\label{sec:vision}
We discuss here our vision for an ecosystem of the users, infrastructure, and interfaces involved in serving \tpms{}.
This vision is guided by our own experience in applying large-scale models to scientific discovery, current research in large-scale model serving, and envisioning future requirements or use cases.

\subsection{User Community}

Large-scale models have already proven useful in more domains than we can enumerate.
The scope of AI for science will continue to expand as \tpms{} are made accessible to the scientific community.
Supporting the diverse and widespread user community motivates much of our vision for the future of \tpms{}.

As we have seen with existing large-scale models, building one (or a few) foundational \tpms{} has proven more powerful and cost-effective than building individual domain-specific models.
While the process of training a foundational \tpm{} for science is outside of the scope of this work, we note that a model of this scale will likely be trained on many data modalities encompassing data from thousands of sources.
Users of \tpms{} will want to perform inference on a similarly diverse set of modalities and data sources.

To accommodate a global community of researchers, we envision that a foundation \tpm{} be replicated at many different locations.
For instance, being co-located with scientific experiments with high throughput requirements such as scientific instruments.
We also expect that \tpms{} support a variety of common data modalities (e.g., text, floating point, graphical, surfaces, point clouds) or even arbitrarily structured formats. 
Support for these various data modalities will be necessary to support the various types of scientific data which may be of interest for these models (e.g., metal organic frameworks, atomic orbitals, epigenetic data, beamline data).

\subsection{Serving Providers}

Research computing centers (RCCs), such as those at universities and government laboratories, are well positioned to support the storage, serving, and maintenance of \tpms{}.
RCCs with flagship compute clusters have the compute capabilities and support staff to make \tpms{} accessible to the research community.
In addition, RCCs (1) provide user authentication and resource quotas, (2) are often co-located with other scientific workloads and experiments, and (3) have high-speed network connections to other research institutions and computing facilities (e.g., Internet2 and ESnet).
Thus, we argue the RCCs at institutions, national cyberinfrastructure providers like ACCESS, and Leadership Computing Facilities at Argonne and Oak Ridge National Laboratories are ideal locations to host \tpms{}. 
We envision a few requirements for hosts of \tpms{}.

First, dedicated compute nodes will be provisioned for persistent serving of one or more \tpms{}.
This could be a subset of nodes on a larger GPU cluster or a smaller dedicated serving cluster.
In the case of a dedicated serving cluster, specialized AI accelerators (e.g., Cerebras, GraphCore, or SambaNova) could improve serving throughput compared to the more general purpose clusters hosted at many RCCs.
Second, \tpms{} need be accessible programmatically (i.e., APIs and SDKs) and via web interfaces.
As noted in \autoref{sec:3_landscape}, we expect the usage of \tpms{} to be tightly coupled with scientific applications executing on compute clusters at the same RCC.
Thus, workloads within the RCC need high throughput access to the \tpm, but we also note it is important to enable other access modes.
For example, external web-based access to a \tpm can enable development, demonstrations, and workshops.
Third, RCCs will need to provide model version control~\cite{kandpal2023git} for \tpms{} that evolve over time through periodic retraining or continual learning.
Reproducability is a foundational component of scientific research, but AI models used for scientific research must also improve over time.
Effective version control can ensure both model advancement and reproducability. 
Given the sheer size of \tpms{}, version control systems of these models will require great space efficiency to avoid exhausting storage resources.

\subsection{Inference Modes}
\label{sec:vision:inference}

The needs of domain scientists and computer scientists are diverse. In order to enable innovation and scientific discovery, simple and common interfaces to \tpms{} may not be sufficient. The current paradigm popularized by ChatGPT relies on a web-based interface that allows for the submission of requests to a mostly closed source, cloud-hosted model. However, we expect the needs of science models to vary vastly from those of a chat-bot, and from use case to use case. Open science models will need to support non-experts who want easy and quick access, through to highly equipped research groups with vast resources who want to be able to customize all parts of the model. Here, we expand upon alternative modes of inference serving and position them within the needs of domain science. Systems researchers working on AI for science will need to build new, dynamic infrastructure that enables these use cases.

\begin{figure}
	\centering
	\includegraphics[width=0.9\linewidth]{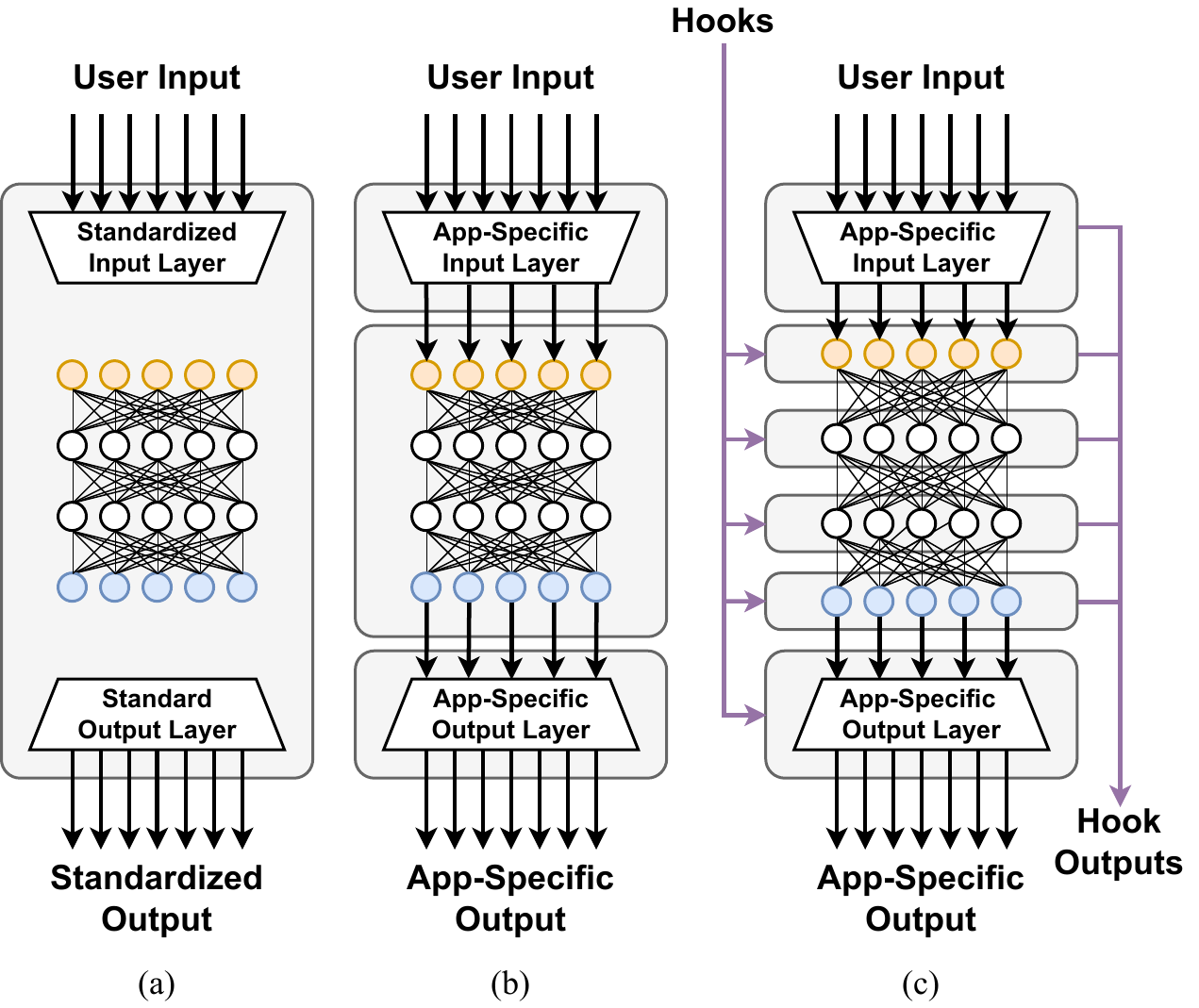}
	\caption{The three envisioned modes of inference. Boxes represent the modular components that go into the respective serving modes; the trapezoids indicate the model portions for ingesting data and generating outputs; and the arrows denote communication between the modules.}
	\label{fig:inference_modes}
\end{figure}

\subsubsection{Query Only}
This use case targets scientists who want to use the foundational model for standardized queries, similar to the model interfaces common today (i.e., chat, sentence completion, etc.). We imagine that this use case could be readily supported by adopting existing approaches to model serving. This approach, seen in Fig.~\ref{fig:inference_modes}(a), allows for the greatest level of optimization and scaling as users interact with the model as a whole rather than customizing input and output or reading and writing intermediary representations. Thus, serving requirements are limited to scaling up and down entire TPM instances.

\subsubsection{Customized Embeddings}
Expert users will want to customize \tpms{} to their own use case. Specifically, large experiments or domains with their own data may want to build custom methods for data ingestion and output on top of the foundational models or even develop additional fine-tuned layers for their own field. We imagine a middle ground approach between using a model service and self-hosting where the core model is the same between all users and each user has their own method for feeding data in and unembedding the outputs from the final core layer---see Fig.~\ref{fig:inference_modes}(b). In this case, the core model acts as a module and the application-specific embedding and unembedding layers are their own modules that can and should be reused across different applications with the same data formatting and output needs.

\subsubsection{Intermediary Accesses}
More advanced users knowledgeable or intrigued by intermediary layer outputs will need a method to access all parts of a model. This would also encompass users who want to introspect the model at different points for things such as machine interpretability~\cite{sakarvadia2023memory}. The challenges with this approach come with designing a flexible serving system that modularizes what is possible while exposing the model via a system of hooks where users can read from and write to specific parts of the model---see Fig.~\ref{fig:inference_modes}(c). In this case, we would want multiple levels of modularity that can be composed dynamically depending on what the user requires. For example, we should be able to independently serve everything from single \texttt{matmul} operations up through entire layers or even blocks of layers. As model weights within the modules are immutable, we are still able to efficiently precompile and scalably serve the portions of the model users do not wish to change while providing the customizability of plugging in custom layers, hooks, or introspection tools.
This is perhaps the least supported by current serving infrastructure.

\section{Challenges}
\label{sec:challenges}
Achieving the vision outlined in~\autoref{sec:vision} will necessitate tackling a wide-range of open challenges.
Theses challenges range from administrative to academic at all levels of the serving stack.
While not exhaustive, this list highlights key areas for innovation.

\subsection{Access Control}
Open access to \tpms{} is necessary to advance scientific progress in a manner which supports the peer review process and removes economic barriers; however, some degree of access control is necessary to enable fair access and prevent misuse of resources.
Open \lsms{} have a very different economic model in contrast to proprietary \lsms{} offered by \cmps{}.
For example, a \cmp{} may charge users per thousand tokens of input or per inference query. 
This profit-driven model enables a \cmp{} to scale resources in or out based on demand.
Comparatively, a RCC hosting an open \tpm{} will be operating in a resource-constrained environment: budgets or grants are fixed and resources are generally static.

RCCs already provide a mechanism, service units (SUs), for fairly sharing constrained resources among the RCC's user community.
Users can apply for an allocation of SUs over a given time period according to their needs.
The SU allocation system ensures that interested parties have access to resources and that bad actors cannot starve others' access.
However, SUs at modern RCCs are typically very coarse grained---equating to a core-hour or GPU-hour.
An allocation system for \tpm{} access will require more granular control.
Naively, an RCC could charge per query, but not all queries are created equal---the amount of data provided as input or generated by a model can incur I/O costs on behalf of the provider.

Further, access control mechanisms that limit what aspects of a model or serving infrastructure users can interact with may be necessary to prevent misuse of resources.
For example, a researcher may want to fine-tune part or all of a \tpm{}.
This may be prohibitively expensive in terms of compute, storage, and I/O, so providers may need to limit the degree to which this can be done.
Further, a custom fine-tuned model may need additional access control to avoid leaking sensitive data, or worse, to avoid the creation of malicious models.

\subsection{Resource Availability}
Allocations can enable fair access in resource constrained settings, but computing facilities will also need to set and meet service level objectives (SLO).
SLOs in the context of \tpm{} serving may include average response time, maximum queries per second, or uptime.
Scientists running large-scale workflows that query a \tpm{} will need to know what performance they can rely on when configuring the parameters and scales of individual runs.
If the serving of a \tpm{} is too unreliable, such as exhibiting high variance in response times, then subsequent programs interacting with the \tpm{} will exhibit the same inconsistencies.
This could be costly in terms of machine utilization for high-performance computing applications.

It may also be necessary to provide tiers connected to the SUs a user is allocated.
A users wanting to make millions of queries to a \tpm{} for a data processing task has very different performance requirements from a user that wants to hook into the intermediate layers of a model to interpret a single model output in fine detail.
Similar to \cmps{} like OpenAI which serve multiple kinds of models at different cost tiers, different resource tiers based on use case can enable providers to better optimize deployments and independently scale different services in or out.

Another avenue for consideration is how users can bring their own allocations or resources to host the model.
For example, can a project with an existing allocation on a GPU cluster use that cluster to spin up an instance of the \tpm{} for their sole ephemeral use while conducting a large-scale experiment?
Or would this project need to request the the provider scale out the \tpm{} serving ahead of time for their experiment? Services such as Globus Compute~\cite{chard20funcx}---a federated Function-as-a-Service platform---may provide a basis for enabling a ``bring your own'' compute model.

\subsection{Model Evolution}

AI advances so rapidly that it is necessary to consider how quickly current \tpms{} will become obsolete.
Organizations may be unwilling to make temporal or financial investments in developing \tpm{} serving infrastructure that quickly becomes deprecated.
System infrastructure will need to be adaptable to future models as the goals and model architectures change over time.

Beyond the hazy horizon of future model architectures, an immediate challenge is that of maintaining a single model that evolves over time.
A model can evolve because it is being continually trained or periodically retrained
(we discuss the technical aspects of updating models in more detail in~\autoref{sec:challenges:updating:retraining}).
Updating a \tpm{} designed to enable science is particularly pertinent because scientific knowledge is continually produced and updated.
Thus, there will be a trade-off between keeping \tpms{} up-to-date and reproducible.
For example, the storage costs would be prohibitive to store a complete copy a \tpm{}'s weights when the model is updated on a regular basis (e.g., daily through continual training).
A robust, scalable, and storage-efficient versioning system is necessary to reconcile the storage requirements with the need for researchers to be able to reproduce their results using prior model versions.

\subsection{Interfaces}
\label{sec:vision:interfaces}

To make a foundational \tpm{} accessible to a wide range of users, including those with limited technical knowledge, several essential aspects must be considered. First, the user interface should be intuitive and user-friendly, ensuring that users can interact with the AI model effortlessly. Moreover, providing ``multilingual'' support or API translations can make the AI model accessible to users from diverse programming language backgrounds. Customization of output is another important aspect, allowing users to adjust technical depth, response format (text, audio, visual), and content to suit their specific needs and use cases. Simplifying the model embedding extraction process makes it easier for users to integrate this information into their own workflows. Additionally, comprehensive and user-friendly documentation and tutorials should be developed to cater to users with varying levels of technical expertise. Encouraging community involvement by allowing users to share their customizations, templates, and accessibility improvements can enhance usability and accessibility. Finally, ensuring scalability is essential to accommodate a growing user base without compromising performance and accessibility.

Researchers will need access to interfaces which enable more fine grain access during a forward pass.
Current large cloud-hosted LLMs typically only allow users access to the final embedding object or the decoded output.
Customization of the forward pass is typically limited to parameters which modify the final decoded output such as the response format, number of unique responses, and temperature for controlling randomness.
These APIs will need to be extended to support open model research, but it is also not possible to know of all manners of access that will be required.
Thus, it may be necessary to allow users to hook into the model's forward pass.
Deep learning libraries already provide mechanisms for registering hooks onto component modules of a model (e.g., PyTorch's \texttt{register\_module\_forward\_hook()}).
Careful consideration must be given to the design of a feature like this due to security concerns, but this is a powerful and novel feature which can enable far more advanced research into \tpm{}s. 

\subsection{Interpretability}
\label{sec:transparency}

\begin{figure}
    \centering
    \includegraphics[width=0.8\linewidth]{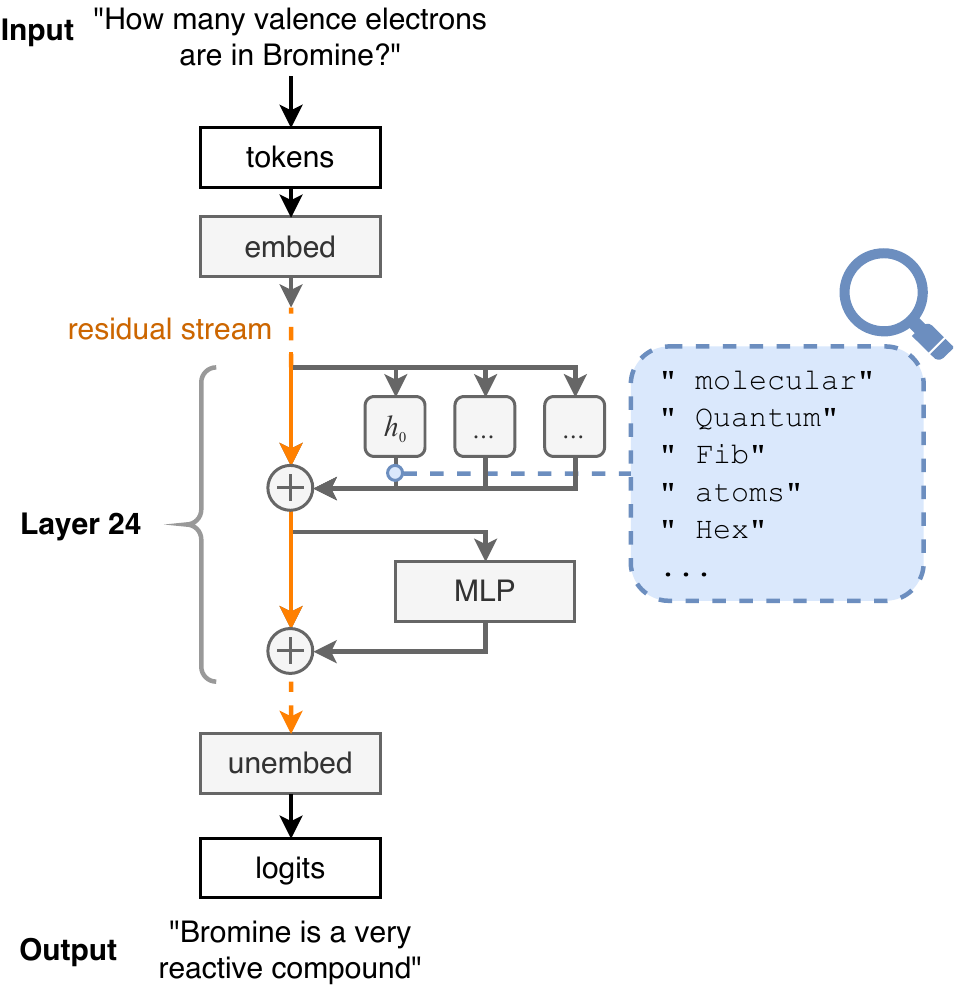}
    \caption{
        Interpreting the \textit{residual stream} with a lens framework that can directly give intermediate outputs from individual attention heads in GPT-like LLMs. This figure uses input, intermediate, and output values from \texttt{GPT2-Large}.
    }
    \label{fig:resid_stream}
\end{figure}

As ML models are increasingly used in scientific computing, it is important to understand what spurious algorithmic behavior may emerge and the degree to which there is uncertainty in a calculation. However, extracting such behavior/metrics from many ML workflows is an open research problem. 
\tpms{} for science need increased transparency to allow researchers to investigate their properties and behaviours (some of which may not be known). To elevate transparency it is important to make public the training/validation dataset, the training procedure details, and model weights. However, in many cases, due to massive scale or intellectual property concerns, it may not be possible to make available one or more of these pieces of information.

For these reasons, it is important to provide as many model probing capabilities as possible when serving a \tpm{}. This need is validated by recent work in ML interpretability which attempts to reverse engineer how behaviors emerge in models, typically by compiling empirical evidence for behavior mechanisms by causally probing a model~\cite{wang2022interpretability,nanda2023progress,geva2023dissecting}. A critical tool in these interpretability workflows is the ability to \textit{hook} the output of intermediate layers to understand what the model may be \enquote{thinking} as it refines the final output~\cite{tuned_lens, logit_lens, sakarvadia2023attention}.
For example, in \autoref{fig:resid_stream}, researchers directly investigate what tokens (or parts of known vocabulary) the model is paying attention to during the forward pass.
Current research in mechanistic interpretability of pre-trained language models have investigated how to directly edit the weights of these models to correct disinformation without needing to retrain them~\cite{ROME}. 
These techniques, as they mature, may be an attractive alternative to correcting a science-oriented \tpm{} without need to perform costly retraining.
Recent interpretability research has investigated how to meaningfully \enquote{nudge} models towards more correct reasoning during the forward pass using injection mechanisms~\cite{sakarvadia2023memory}. Similar methods can be developed to bootstrap scientific idea generative tasks to aid interdisciplinary research and discover new connections across scientific domains.

Enabling this interpretability and transparency research with \tpms{} will influence API design considerations as mentioned in~\autoref{sec:vision:interfaces}, and we discuss the performance implications of saving and returning intermediate layer outputs in~\autoref{sec:challenges:inference}.

\subsection{Incorporating New Knowledge}
\label{sec:challenges:updating:retraining}
In a world where data is being created from single sources at more than a PB/day~\cite{doe2022aiworkshop},
updating models to incorporate new insights and knowledge becomes a non-trivial challenge. Two primary solutions that currently exist for this problem are vector databases---to enable models able to encode and subsequently query keys to access data sources---and retraining, which is prohibitively slow at this scale.

\subsubsection{Vector Databases}
Popular commercial solutions such as Pinecone~\cite{pinecone} enable end users to create stores of data (or information) from different sources that makes them queriable and accessible by common model architectures. This type of technology could potentially be scaled up to handle multiple simultaneous encoding workflows (e.g., defining how to handle data at the observation, experiment, or data level). Additionally, such a system would have to incorporate static embedding methods across all queriable knowledge accessible to the \tpm{} such that all models can retrieve from the same data store. This consideration would require some static connotation across all applications of the \tpm{}, which could be suboptimal for certain fine-tuned science applications. However, the inverse could also be beneficial in data siloing via unique query/embedding spaces.

\subsubsection{Retraining}
Accommodating novel data can also include selective retraining of the \tpm{}. The tradeoffs and forms this may present as are many and complex. For example, the principles outlined in ~\autoref{sec:transparency}
could incorporate or edit knowledge identifiable within the model. Additionally, some sort of selective threshholding can be used to update the model based on the amount or uniqueness of data collected. Quantifying the tradeoff is extremely important as any sort of whole-model retraining is subject not only to extremely high compute costs but also the potential of degrading or altering response characteristics for other tasks.

\subsection{Integration into Scientific Workflows}
The ultimate success or failure of \tpms{} in science will depend on the scale and speed of adoption of these methods by science applications. The primary barrier to widespread adoption is integrating existing science interfaces (e.g., workflows, data, control processes, etc.) with interfaces to foundational AI infrastructure. Here we discuss the challenges that must be overcome to facilitate that alignment.

\subsubsection{Application types}
The first type of application is general use; the type of integration we have seen so far in uses of commercial infrastructure (e.g., ChatGPT, Bard) in software and science workflows. On the science side, these uses involve transforming the science query into the format the model was trained to accept. For an LLM, this requires translating a scientific query into a human readable prompt. Expressing highly technical information (e.g., molecular diagrams) as plain text can be challenging. Existing translation methods have merit, such SMILES strings for molecules, but model confusion can be common if training data lacks sufficient examples of a given representation. Multi-shot inference methods resolve some of these challenges, but this often requires significant human curation which can be a barrier to adoption.

The second application relates to using the core of foundational models with custom data ingestion techniques. In such a case, unique data can be used with existing models while changing only ingestion and output modalities (described further in \autoref{sec:vision:inference}). For instance, a common example is the use of traditional CNNs to embed images for use by LLMs. 
We envision this being a prime use case for science as scientists do not want to and should not have to significantly transform their existing data and workflows to work within a specific model modality. In this case, users could use and contribute embedding methods for different data formats and in return get access to a wide range of domain embedding methods. Think of this as a model zoo for AI data ingestions. Correspondingly, we see parallels with model outputs as different science use cases will require very different end states, particularly given the contrast between currently prevailing autoregressive methods and more traditional methods like classification tasks.

The third set of applications require direct access to internal model representations as well as the ability to inject altered or transformed versions of these representations. Such applications will require the model to be modular enough to break apart into interactive elements while allowing the unperturbed portions to operate independently and efficiently. These use cases could include work in interpretability, interactive AI environments, and deeper research into next-gen large models. The challenge here lies in the interface to end-users. There will need to be some method to enable programmatic and interpretable access to the inner portions of the model, requiring on-the-fly recompilation. Additionally, any loading or alteration of the model will have to rely on the aforementioned modularity as it is infeasible to expect tests altering a small portion of the model to require reloads or recompilations of the entire model.

\subsubsection{Interfaces}
While the two most commonly seen interfaces in current model serving offerings, namely text-based GUIs and APIs, will undoubtedly be necessary (and likely most important) for serving \tpms{}, additional interfaces will undoubtedly be required. The two primary examples of additional interfaces are an interface to interact with inner portion of the model (discussed in greater length in the previous section and in \autoref{sec:vision:inference}) and an interface to enable model access to data repositories and data streams. The challenge of internal interaction (beyond the previously discussed serving and modularity systems challenges) lies primarily in defining how scientists would need to access weights and intermediary representations. One critical roadblock is when intermediary outputs and model weights need to be observable and editable simultaneously. For example, the challenge of applying some normalization across all the model weights and observing the effects on output would be nontrivial. To interface models with data, there will need to be a mechanism developed to allow for multiple inputs (i.e., the data and the user query), that is plug-and-play and does not require alteration of existing science infrastructure.

\subsubsection{Tradeoff with usability and customizability}
Overall, these different levels of serving and interoperability demonstrate a new tradeoff between model usability and customizability. One middle ground is to create a model zoo for the embedding and unembedding/output layers to enable reuse for a given problem or data type within a domain. By creating a model zoo like this for different data types and applications, we have the ability to create and serve better optimized versions of these smaller model modules given they will be reused within some domain of application and data. Additionally, this can enable domain science engagement by offering an embedding creation service where scientists receive a fine-tuned embedding/encoding to connect their workflow to the core model and then the hosting team gets access to troves of new data and increasingly diverse application areas. 

The challenge lies in enabling and automating such a service and effectively serving the outputs. Initially, there will be overheads of offering this in a centralized manner. However, automating the service over time would offer greater benefits in terms of efficiency with the slight cost of less direct end-user feedback. Enabling this type of service gives access to more data which can be used to further improve the abilities of the core model. Combined with the automated embedding/integration service, automating updates of the core model could form a feedback look similar to AutoML~\cite{automl}---by better ingesting data, we can access more and better quality data to improve the core model, which will then enable a wider range of use cases and so on.

\subsection{Efficient Inference}
\label{sec:challenges:inference}
Supporting the variety of inference modes outlined in~\autoref{sec:vision:inference} and~\autoref{sec:vision:interfaces} raises two key questions: does flexibility in model usage come at the cost of inference efficiency and what does an ideal interface look like for flexible inference patterns.
In contrast to for-profit AI services which are motivated by decreasing cost-per-query and therefore increasing profit margins, hosting open models may necessitate relaxing efficiency goals.
Nevertheless, providing flexible access to open models must still be done in a manner that maintains high system utilization to (1) balance costs with academic progress and (2) support more users.

Efficient inference with a \tpm{} requires optimization at several levels in the stack. 
At the highest abstraction level, the model architecture, layer and weight partitioning across time (multiplexing) and space (the compute cluster) enables parallelism and concurrency such as in DeepSpeed~\cite{rajbhandari2020zero,aminabadi2022deepseedinference}, Megatron-LM~\cite{narayanan2021megatron}, or PyTorch's Fully Sharded Data Parallel~\cite{paszke2019pytorch}.
Optimally solving this scheduling problem is challenging due to the heterogeneity of inter-GPU communication in a multi-node GPU cluster
(i.e., the latency and bandwidth between two GPUs in the same node is different than between two GPUs in different nodes or even different racks).
Performance is difficult to model accurately so typically profiling is done---either manually or automatically as in Alpa~\cite{zheng2022alpa}.

At a lower level, kernel fusions are crucial for efficient computation on a single compute device.
Kernel fusions lead to improved arithmetic intensity (i.e., data reuse), which is reflective of better cache locality and potential use of various fused hardware intrinsics---fused multiply-accumulate (FMA) and matrix multiply and accumulate (MMA).
Such techniques generally require a global view of the model (i.e., the whole computational graph) and are thus at odds with distributed inference techniques which partition the model.
Fusions are primarily compiler driven---transparent to the user---but can be helped or harmed by user choices.

Last, at the system level of abstraction (i.e., irrespective of the unique properties of \lsms{}) efficient memory allocation can affect performance.
Assuming that any inference environment is multi-tenant, concurrent threads can and will make contending demands on the system's memory allocator.
A typical high-performance allocator will be a multithreaded arena allocator, reserving a small scratch space on a per-thread basis and a single global arena for large allocations.
The boundary (allocation size) between these two allocation ``regimes'' is often user-configurable, at both the system and application level, and can affect performance through lock contention.

With this perspective on the complexities of optimizing inference in mind, consider, for example, an inference use case which needs access to the intermediate embedding outputs of the transformer blocks in a large transformer (e.g., 100s of transformer blocks) such as for knowledge-editing (\autoref{sec:transparency}).
Inference on large models can be optimized because the computational graph is typically static which allows for sequential operations to be fused together, pre-allocation of memory, and more, but these techniques could be inhibited if intermediate values need to be returned to the user dynamically. 
Total memory usage would vary from inference batch to inference batch, and tensor memory allocations will no longer be static.
Immediately moving intermediate values from accelerator memory to host memory could alleviate some of these problems, but would introduce memory copy and synchronization overheads.
Thus, while there exists mechanisms for optimizing inference serving, further work needs to be done to support the more flexible deployment scenarios required by the scientific community.

\section{Conclusions}
\label{sec:conclusions}
We have discussed the challenges and opportunities that lie ahead as the scientific community embraces \lsms{} as a tool for scientific discovery.
We have presented a vision for an ecosystem that bridges the gap between TPM providers and users, with a focus on enabling researchers to address novel and large-scale problems.
Moreover, our exploration has highlighted the technical challenges and open problems associated with serving TPMs, emphasizing the need for investments and a comprehensive software stack that can accommodate the diverse requirements of researchers.
In the coming years, the collaboration between AI researchers, developers, and the scientific community will be crucial in addressing the complexities of TPMs and ensuring that these models become valuable tools for advancing knowledge and innovation.

\begin{acks}
This work was supported in part by the U.S.\ Department of Energy under Contract DE-AC02-06CH11357.
\end{acks}

\balance
\bibliographystyle{ACM-Reference-Format}


\end{document}